\documentclass{article}

\newif\ifarXiv
\arXivfalse
\arXivtrue
\ifarXiv
    \usepackage[nonatbib, preprint]{neurips_2021}
\else
  \usepackage[nonatbib]{neurips_2021}
\fi
\usepackage[utf8]{inputenc} 
\usepackage[T1]{fontenc}    
\usepackage[hidelinks]{hyperref}       
\usepackage{url}            
\usepackage{booktabs}       
\usepackage{amsfonts}       
\usepackage{nicefrac}       
\usepackage{microtype}      
\usepackage{graphicx}
\usepackage{comment}

\usepackage{amsmath}
\usepackage{tikz}
\usetikzlibrary{shapes}

\title{Targeted Deep Learning:\\
Framework, Methods, and Applications}

\author{%
  Shih-Ting Huang \\
  Department of Mathematics\\
  Ruhr University, Bochum\\
  \texttt{shih-ting.huang@rub.de} \\
  \And
  Johannes Lederer \\
  Department of Mathematics\\
  Ruhr University, Bochum\\
  \texttt{johannes.lederer@rub.de} \\
}
\usepackage{xcolor}

\newcommand{\eqv}{:= }
\usepackage{amssymb, amsmath}
\begin{document}

\maketitle

\begin{abstract}
Deep learning systems are typically designed to perform for a wide range of test inputs.
For example, deep learning systems in autonomous cars are supposed to deal with traffic situations for which they were not specifically trained.
In general, the ability to cope with a broad spectrum of unseen test inputs is called generalization.
Generalization is definitely important in applications where the possible test inputs are known but plentiful or simply unknown,
but there are also cases where the possible inputs are few and unlabeled but known beforehand.
For example,
medicine is currently interested in targeting treatments to individual patients;
the number of patients at any given time is usually small (typically one),
their diagnoses/responses/... are still unknown,
but their general characteristics (such as genome information, protein levels in the blood, and so forth) are known before the treatment.
We propose to call deep learning in such applications $\emph{targeted deep learning.}$ 
In this paper, we introduce a framework for targeted deep learning,
and we devise and test an approach for adapting standard pipelines to the requirements of targeted deep learning.
The approach is very general yet easy to use: 
it can be implemented as a simple data-preprocessing step.
We demonstrate on a variety of real-world data that our approach can indeed render standard deep learning faster and more accurate when the test inputs are known beforehand.
\end{abstract}

\section{Introduction}
Deep learning pipelines are typically designed to generalize, that is, to perform well on new data.
Consider pairs of inputs and outputs $(\boldsymbol{x}_1,y_1),\dots, (\boldsymbol{x}_n,y_n)\in\mathbb{R}^p\times\mathbb{R}$ and a corresponding loss function $\ell\,:\,\mathbb{R}\times\mathbb{R}\to[0,\infty)$.
In practice,
 generalization is typically measured through sample splitting: 
 after calibrating the parameters of the pipeline on a training set $\mathcal{T}\subset\{1,\dots,n\}$,
 the corresponding network~$\mathfrak{g}$ is tested on the validation set $\mathcal{V}\eqv\{1,\dots,n\}\setminus\mathcal{T}$:
\begin{equation*}
    \text{empirical generalization error}~\eqv~\sum_{i\in\mathcal{V}}\ell\bigl[\mathfrak{g}[\boldsymbol{x}_i],y_i\bigr]\,.
\end{equation*}
In theory,
generalization is typically measured through statistical risks~\cite{bartlett_mendelson_2001,lederer2020risk}:
\begin{equation*}
    \text{theoretical generalization error}~\eqv~\mathbb{E}_{(\boldsymbol{w},v)}\,\ell\bigl[\mathfrak{g}[\boldsymbol{w}],v\bigr]\,,
\end{equation*}
where $\mathbb{E}_{(\boldsymbol{w},v)}$ is the expectation over a random sample $(\boldsymbol{w},v)\in\mathbb{R}^p\times\mathbb{R}$.
Both empirical and theoretical generalization errors essentially describe a network's average error on new data.

The strength of generalization is that it captures, or at least tries to capture, a network's performance on \emph{unknown} inputs.
For example, generalization seems appropriate for training fraud-detection systems,
which might face all sorts of unforseen input~$\boldsymbol{w}$.
But in some cases,
the relevant inputs are already \emph{known}.
For example,
assume that we have covariate vectors~$\boldsymbol{x}_i$ and corresponding diagnoses~$y_i$ for $n$~patients,
and that we want to use these data to diagnose a new group of patients with known but unlabeled covariates~$\boldsymbol{w}_1,\dots,\boldsymbol{w}_g$.
We call $\boldsymbol{w}_1,\dots,\boldsymbol{w}_g$ the \emph{target vectors}.
We could, of course, still train a deep learning pipeline towards generalization as usual,
but this does not seem optimal at all,
because it completely disregards the information about the unlabeled target vectors~$\boldsymbol{w}_1,\dots,\boldsymbol{w}_g$. 
Our objective in this paper is, therefore, to train networks for optimal performance for individual or groups of known but unlabeled target vectors.
For further reference,
we call this objective \emph{targeted deep learning}.

The idea of including target  vectors into data-driven decision-making processes has already gained track recently in medical statistics under the names ``precision medicine'' and ``personalized medicine''~\cite{harvey_brand_holgate_kristiansen_lehrach_palotie_prainsack_2012, huang2019tuning,hellton2020penalized}.
A main motivation for this paper is to highlight the potential of this general idea in deep learning.
Moreover, 
we devise a concrete approach for deep learning that is suprisingly simple and yet can be used in a wide variety of frameworks.
Rather than changing the pipelines themselves,
as done previously in medical statistics,
we include the targeting into the optimization routines.
We finally illustrate our framework and methods in applications in medicine, engineering, and other fields.

Our three main contributions are:
\begin{itemize}
    \item We establish the concept of targeted deep learning.
    \item We device a simple yet effective approach to targeted deep learning.
    \item We highlight the benefits of prediction deep learning in practice.
\end{itemize}

\paragraph{Outline}
In Section~\ref{sec:frameworkmethods},
we develop a statistical framework and a concrete algorithm for targeted deep learning.
In Section~\ref{sec:applications},
we show the effectiveness of our approach on real-world data.
In Section~\ref{sec:discussion},
we highlight the potential impact of our paper on precision/personalized medicine and beyond.

\paragraph{Related work}
The framework of targeted deep learning as defined above has not been studied in deep learning before.
But other frameworks that deviate from the standard setup of generalization have been studied already.
We briefly discuss some of in the following.
The arguably most related one is \emph{personalized deep learning}~\cite{rudovic_lee_dai_schuller_picard_2018,schneider2020personalization}.
Personalized deep learning assumes that each sample corresponds to one of~$m$ separate entities (such as patients, customers, countries, ...).
The goal is then to train  separate networks for each of these entities.
There is a range of  approaches to personalized deep learning:
Networks are trained on the whole dataset first and then trained further on the subset of data that corresponds to the entity of interest;
this approach is called \emph{transfer learning}~\cite{transfer1993, zhuang2020comprehensive}.
Alternatively, networks are first trained on the samples of the entity of interest and then trained further on the entire data;
this approach is called \emph{early shaping}~\cite{schneider2020personalization,pmlr-v28-sutskever13}.
Or, instead, networks are trained on the samples of the entity of interest  combined with similar samples from the entire dataset;
this approach is called \emph{data grouping}~\cite{schneider2020personalization,shorten_khoshgoftaar_2019}.
Networks can also be trained on features specifically designed for personalization~\cite{gupta2017learning}.
In addition,
there are approaches designed for specific tasks, such as tag recommendation~\cite{nguyen_wistuba_grabocka_drumond_schmidt-thieme_2017}.
While personalized deep learning and targeted deep learning can be used on similar types of data,
they differ substantially their goals and requirements:
personalized deep learning seeks  optimal generalization for a given entity (``person''),
while target deep learning seeks optimal prediction for an input or a group of inputs---whether these inputs belong to a common entity or not;
personalized deep learning requires (a sufficiently large number) of labeled samples for the entity at hand,
while targeted deep learning  works even for single, unlabeled target inputs.
Hence, none of these mentioned methods applies to our framework.

But then there are also other, less related frameworks.
One, very well-known framework is  \emph{unsupervised learning} (such as \emph{autoencoders}~\cite{kramer_1991}),
where
labels are missing altogether,
and where the goal is not prediction but instead a useful representation of the inputs.
There is also supervised learning with \emph{unlabeled data in the presence of domain, knowledge},
where labels are missing but replaced by a constraint functions,
and where the goal is generalization~\cite{stewart2016labelfree}.
Another framework is \emph{multimodal learning},
where data from different sources is used~\cite{8103116}.
One more is \emph{adaptive learning},
where the algorithm can query a label of a previously unlabeled sample~\cite{cortes2017adanet}.

Targeted deep learning does not compete with these frameworks and methods:
our concepts and methods simply provide an approach to a type of application that has not yet been considered in deep learning.

\section{Framework and methods}
\label{sec:frameworkmethods}

We first establish a framework for targeted deep learning and contrast targeted deep learning with precision deep learning.
We then devise a general yet very simple approach to gear standard deep learning pipelines toward targeted deep learning.

\newcommand{\drawmyplot}{
\scalebox{0.7}{\begin{tikzpicture}
\newcommand{\sizeval}{6}
\newcommand{\lineval}{1.1}
\node[draw, cloud, cloud puffs=10, cloud puff arc=120, aspect=3, line width=\lineval] (model) at (0, 0) {deep learning model};

\node[draw, text width=30mm, align=center, line width=\lineval] (inputdata) at (-\sizeval, 0) {labeled~training~data\\
$(\boldsymbol{x}_1,y_1)$\\
$(\boldsymbol{x}_2,y_2)$\\
$\cdots$};
\node[draw, text width=30mm, align=center, line width=\lineval] (targets) at (\sizeval, 0) {unlabeled test data\\
$(\boldsymbol{w}_1,?)$\\
$(\boldsymbol{w}_2,?)$\\
$\cdots$};
\node[red, draw, text width=30mm, align=center, , line width=\lineval] (targets2) at (-\sizeval, -2) {unlabeled test data\\
$(\boldsymbol{w}_1,?)$\\
$(\boldsymbol{w}_2,?)$\\
$\cdots$};

\node (trainlow) at (-3.25, 0) {};

\draw[->, line width=\lineval] (inputdata) -- (model);
\draw[->, line width=\lineval] (model) -- (targets);
\draw[->, red, line width=\lineval] (targets2) -| (trainlow);

\node (train) at (-3.25, 0.22) {train};
\node (apply) at (3.25, 0.22) {apply to};
\end{tikzpicture}}
}
\newcommand{\drawmyfigure}[1]{\begin{figure}[#1]
    \centering
    \drawmyplot
    \caption{
    our methods adjust standard deep learning (black parts) for targeted deep learning by using the unlabeled test data (red box) to focus the training algorithm
    }
    \label{fig:overview}
\end{figure}
}
\subsection{A framework for targeted deep learning}
\label{sec:framework}

Consider an arbitrary (non-empty) class of neural networks $\mathcal{G}\eqv\{\mathfrak{g}_{\boldsymbol{\theta}}:\mathbb{R}^p\to\mathbb{R}~\text{with}~\boldsymbol{\theta}\in\Theta\}$ parameterized by $\Theta\subset\mathbb{R}^q$ and training data $(\boldsymbol{x}_1,y_1),\dots ,(\boldsymbol{x}_n,y_n)\in\mathbb{R}^p\times\mathbb{R}$.
The usual goal of deep learning is to find a $\widehat{\boldsymbol{\theta}}\equiv\widehat{\boldsymbol{\theta}}[(\boldsymbol{x}_1,y_1),\dots, (\boldsymbol{x}_n,y_n)]$ to achieve a low risk or generalization error:
\begin{equation*}
    \mathbb{E}_{(\boldsymbol{w},v)}\bigl[\ell[\mathfrak{g}_{\boldsymbol{\theta}}[\boldsymbol{w}],v]\bigr]
\end{equation*}
where $\mathbb{E}_{(\boldsymbol{w},v)}$ is the expectation over a random sample $(\boldsymbol{w},v)\in\mathbb{R}^p\times\mathbb{R}$ that describes the training data, and $\ell:\mathbb{R}\times\mathbb{R}\to[0,\infty)$ is a loss function.
This goal is called \emph{generalization}.

Recently,
there has also been research on targeting the performances to entities that correspond to a subset of the data.
In other words,
one wants to maximize the performance of the deep learning pipeline for an entity (patient, client, ...) that corresponds---without loss of generality---to the first~$m$ training samples $\{(\boldsymbol{x}_1,y_1),\dots, (\boldsymbol{x}_m,y_m)\}$.
The goal is then to find a parameter $\widehat{\boldsymbol{\theta}}\equiv\widehat{\boldsymbol{\theta}}[(\boldsymbol{x}_1,y_1),\dots, (\boldsymbol{x}_n,y_n)]$ (which can depend on the entire data set) that generalizes well for that individual entity:
\begin{equation*}
    \mathbb{E}_{(\boldsymbol{w}^{\operatorname{ent}},v^{\operatorname{ent}})}\bigl[\ell[\mathfrak{g}_{\boldsymbol{\theta}}[\boldsymbol{w}^{\operatorname{ent}}],v^{\operatorname{ent}}]\bigr]
\end{equation*}
where $\mathbb{E}_{(\boldsymbol{w}^{\operatorname{ent}},v^{\operatorname{ent}})}$ is the expectation over a random sample $(\boldsymbol{w}^{\operatorname{ent}},v^{\operatorname{ent}})\in\mathbb{R}^p\times\mathbb{R}$ that describes the distribution of the entity in question, that is, the first $m$~training samples (rather than all samples).
Of course, if all samples are independent and identically distributed,
it holds that $\mathbb{E}_{(\boldsymbol{w},v)}[\ell[\mathfrak{g}_{\boldsymbol{\theta}}[\boldsymbol{w}],v]]=\mathbb{E}_{(\boldsymbol{w}^{\operatorname{ent}},v^{\operatorname{ent}})}[\ell[\mathfrak{g}_{\boldsymbol{\theta}}[\boldsymbol{w}^{\operatorname{ent}}],v^{\operatorname{ent}}]]$,
but the basic notion here is that the  individual entity can very well be different from the others.
The number of training samples~$m$ is usually considered much smaller than the total number of samples~$n$ (otherwise, there would hardly be a difference to the usual framework of deep learning),
but $m$ is also assumed to be much greater than~$1$ (because the known approaches involves steps that only have the $m$ first samples at disposal): $1\ll m\ll n$.
The described framework is called \emph{personalized deep learning}.
A generic application of personalized deep learning are recommender systems~\cite{aggarwal_2018}.

In other applications, however, we have both training data and a (potentially very small) group of unlabeled target inputs. 
Our goal is then to design a deep learning pipeline that works well for those targets.
Mathematically speaking,  
we assume given a group of $g$ target vectors~$\boldsymbol{w}_1,\dots,\boldsymbol{w}_g$ and try to find a parameter $\widehat{\boldsymbol{\theta}}\equiv\widehat{\boldsymbol{\theta}}[(\boldsymbol{x}_1,y_1),\dots, (\boldsymbol{x}_n,y_n),\boldsymbol{w}_1,\dots,\boldsymbol{w}_g]$ that has a low empirical loss over these targets:
\begin{equation*}
    \sum_{i=1}^g\ell[\mathfrak{g}_{\boldsymbol{\theta}}[\boldsymbol{w}_i],v_i]\,,
\end{equation*}
where $v_1,\dots,v_g\in\mathbb{R}$ are the unknown labels of the target vectors.
We call this framework \emph{targeted deep learning}.
A generic application is precision medicine,
where we attempt to select a diagnosis/treatment/... for an individual patient that has not been diagnosed/treated/... beforehand.

We see in particular that personalized deep learning and precision deep learning are quite different:
 personalized deep learning is about heterogeneity in the data,
while precision deep learning is about targeting the learning to prespecified inputs.
More broadly speaking,
precision deep learning provides a framework for tasks not covered by existing frameworks but increasingly relevant in practice---see our Section~\ref{sec:applications}, for example.

\subsection{Methods for targeted deep learning}
\label{sec:methods}

Having specified the goal of targeted deep learning,
we introduce methods for training the parameters accordingly.
As described in Section~\ref{sec:framework}, methods for personalized deep learning require (typically large groups of) labeled data for each individual and have an entirely different goal altogether;
similarly, methods for unsupervised deep learning, multimodal learning, and so forth are not suitable here (compare to our previous section on related literature).
Our methods, motivated by~\cite{bu_lederer_2021,huang2019tuning},
instead treat the target vectors as additional knowledge that can be integrated into the training procedure.
A overview of this concept is provided in Figure~\ref{fig:overview}.
Specifically,
we integrate the target vectors into the optimization algorithm in a way that requires minimal changes to  existing deep-learning pipelines and yet is very effective.

\drawmyfigure{t}

The starting point of our approach is a measure for the similarity between a training sample and the target vectors.
In mathematical terms,
we assume a function
\begin{align*}
    \mathfrak{s}~:~\mathbb{R}^p\times\mathbb{R}^p\times\mathbb{R}^p\times\dots\,&\to\,[0,\infty)\\
    \boldsymbol{x},\boldsymbol{w}_1,\boldsymbol{w}_2,\dots\,&\mapsto\,\mathfrak{s}[\boldsymbol{x},\boldsymbol{w}_1,\boldsymbol{w}_2,\dots]\,.
\end{align*}
The first argument of the function is for the input of the training sample under consideration;
the remaining arguments are for the target vectors.
Formally, the function can have arbitrarily many inputs to account for arbitrarily many target vectors,
but in practice, 
the number of target vectors is typically limited.
In any case,
we call the function~$\mathfrak{s}$ the \emph{similarity measure}.

One can take whatever similarity measure that seems suitable for a specific application.
But it turns out that the ``classical choice'' (see Appendix~\ref{sec:max} for further considerations)
\begin{equation*}
    \mathfrak{s}[\boldsymbol{x},\boldsymbol{w}_1,\boldsymbol{w}_2,\dots]~\eqv~\max\biggl\{0,\frac{\langle\boldsymbol{x},\boldsymbol{w}_1\rangle}{|\!|\boldsymbol{x}|\!|_2|\!|\boldsymbol{w}_1|\!|_2},\frac{\langle\boldsymbol{x},\boldsymbol{w}_2\rangle}{|\!|\boldsymbol{x}|\!|_2|\!|\boldsymbol{w}_2|\!|_2},\dots\biggr\}\,,
\end{equation*}
where $\langle\cdot,\cdot\rangle$ is the standard inner product on~$\mathbb{R}^p$ and $|\!|\cdot|\!|_2$ the Euclidean norm on~$\mathbb{R}^p$,
works for a wide range of applications---see especially our applications in Section~\ref{sec:applications}.
Recall from basic trigonometry that 
\begin{equation*}
 \frac{\langle\boldsymbol{x},\boldsymbol{w}_i\rangle}{|\!|\boldsymbol{x}|\!|_2|\!|\boldsymbol{w}_i|\!|_2}~=~\operatorname{cos}[\measuredangle_{\boldsymbol{x},\boldsymbol{w}_i}]   
\end{equation*}
is the cosine of the angle between the vectors $\boldsymbol{x}$ and $\boldsymbol{w}_i$;
hence, geometrically speaking,
our choice of~$\mathfrak{s}$ uses angles to describe the similarity between an input and the target vectors (the additional $0$ in the maximum is just a threshold for the angles to ensures that the function~$\mathfrak{s}$ is nonnegative).
On the other hand,
we can also think of the elements of~$\boldsymbol{x}$ and~$\boldsymbol{w}_i$ as random variables and, assuming the vectors are centered, 
\begin{equation*}
 \frac{\langle\boldsymbol{x},\boldsymbol{w}_i\rangle}{|\!|\boldsymbol{x}|\!|_2|\!|\boldsymbol{w}_i|\!|_2}~=~\frac{\sum_{m=1}^p x_m(w_i)_m}{\sqrt{\sum_{m=1}^p (x_m)^2}\sqrt{\sum_{m=1}^p ((w_i)_m)^2}}~=~\operatorname{cor}[\boldsymbol{x},\boldsymbol{w}_i]   
\end{equation*}
as their empirical correlation~\cite{stigler_1989};
hence, statistically speaking,
our choice of~$\mathfrak{s}$ uses correlations to describe the similarity of an input and the target vectors.
Since the correlation has a long-standing tradition in measuring similarities~\cite{6355687,london_1895,stigler_1989,kendall_1938,goodman_kruskal_1979},
we can indeed call our~$\mathfrak{s}$ the classical choice.

Most deep learning pipelines optimize the parameters based on stochastic-gradient descent (SGD)~\cite{Bottou98onlinelearning}.
In the usual mini-batch version,
SGD draws $b$~samples  $(\widetilde{\boldsymbol{x}}_1,\widetilde y_1),\dots,(\widetilde{\boldsymbol{x}}_b,\widetilde y_b)$ uniformly at random without replacement from the training data  $(\boldsymbol{x}_1,y_1),\dots, (\boldsymbol{x}_n,y_n)$,
calculates the gradient with respect to the selected batch of data of size~$b$,
and then updates the parameters accordingly.
We propose to use this strategy with a small twist:
instead of drawing samples uniformly at random,
we propose to draw samples in a way that accounts for the similarity between a sample and the targets.
Specifically,
we propose to draw the samples in each mini-batch from the dataset with probabilities
\begin{equation*}
    \mathbb{P}\bigl\{(\widetilde{\boldsymbol{x}}_j,\widetilde y_j)=({\boldsymbol{x}}_i, y_i)\bigr\}~=~\frac{\mathfrak{s}[\boldsymbol{x}_i,\boldsymbol{w}_1,\boldsymbol{w}_2,\dots]}{\sum_{m=1}^n\mathfrak{s}[\boldsymbol{x}_m,\boldsymbol{w}_1,\boldsymbol{w}_2,\dots]}~~~~~~j\in\{1,\dots,b\},i\in\{1,\dots,n\}\,.
\end{equation*}
This means that we weight the drawing mechanism according to the similarities between the data points and the targets:
the more similar a sample is to the targets,
the more often tends to appear in a batch.

We thus propose to replace the standard uniform drawing scheme in stochastic-gradient decent by a weighted drawing scheme. 
In short, we perform a form of \emph{importance sampling}~\cite{importsample}.
The only difference to other stochastic-gradient descent algorithms with weighted drawing schemes is the drawing distribution: 
our distribution directs the training towards the target inputs,
while the distributions applied in other contexts usually aim at fast numerical convergence~\cite{needell2015stochastic}.

The drawing scheme can, of course, be implemented very easily,
and the computational complexity of a single gradient update is the same as in the original setup.
But to avoid adjusting the code of every deep learning pipeline that one wants to use for targeted deep learning,
or simply for testing reasons,
we can also proceed slightly differently:
instead of altering SGD,
we can simply alter the data it operates on.
More precisely,
we can  generate a new dataset with samples  $(\boldsymbol{x}'_1,y_1'),\dots, (\boldsymbol{x}_{\lfloor t\cdot n\rfloor}',y_{\lfloor t\cdot n\rfloor}')\in\mathbb{R}^p\times\mathbb{R}$ by sampling from the original data with the following probabilities:
\begin{equation*}
    \mathbb{P}\bigl\{(\boldsymbol{x}'_j, y_j')=(\boldsymbol{x}_i, y_i)\bigr\}~=~\frac{\mathfrak{s}[\boldsymbol{x}_i,\boldsymbol{w}_1,\boldsymbol{w}_2,\dots]}{\sum_{m=1}^n\mathfrak{s}[\boldsymbol{x}_m,\boldsymbol{w}_1,\boldsymbol{w}_2,\dots]}\,.
\end{equation*}
The parameter~$t\in(0,\infty)$ scales the number of samples of the new dataset;
in our experience, it is of very minor importance;
for example, $t=10$ works for all practical purposes---see Appendix~\ref{sec:additional}.
In any case, using the standard batching scheme (with uniform sampling) on the new dataset $\{(\boldsymbol{x}'_1,y_1'),\dots, (\boldsymbol{x}_{\lfloor t\cdot n\rfloor}',y_{\lfloor t\cdot n\rfloor}')\}$ then yields the desired distribution of the samples in each batch. 
Thus, rather than changing the implementation of SGD,
we can apply the original pipeline to the new dataset.

While we allow for $t>1$, 
 the purpose of our data-preprocessing scheme is  to simplify the batching---not to augment the data.
In particular,
in contrast to data-augmentation schemes designed for deep learning in data-scarse applications~\cite{9186097},
we do not generate new, artificial samples.

In conclusion,
our approach to targeted deep learning boils down to a simple data-preprocessing step.
This means that 
adjusting existing deep learning pipelines and implementations to targeted deep learning is extremely easy.

\section{Applications}
\label{sec:applications}
We now show that targeted deep learning trains networks faster and more accurately than standard deep learning when given individual or groups of targets.
We consider six applications,
two regressions,
two general classifications,
and two image classifications.
We first explain the general setup, then detail all applications,
and finally provide and discuss the results.

\paragraph{General setup}
Our goal is to compare targeted deep learning with standard deep learning in a way that might generalize to other applications;
in particular,
excessive tailoring of the deep learning setup to the specific data as well as  artefacts from specific deep learning supplements,
such as dropouts~\cite{hinton2012improving}, special activation functions~\cite{lederer2021activation}, complex layering, and so forth,
should be avoided.
More generally,
our interest is not to achieve optimal performance on a given data set (by fine-tuning the system, for example)
but to show that targeted deep learning can improve standard deep learning---irrespective of the specific pipeline---when target vectors are known beforehand.

We focus on small groups ($g=1$ and $g=5$),
because this is the most realistic scenario in the majority of the discussed applications.
However, we show in Appendix~\ref{sec:group} that our approach also works for much larger groups.

We use standard stochastic-gradient decent with constant learning rate~0.005.
Targeted deep learning is implemented as described in Section~\ref{sec:methods}.
The deep learning ecosystem is PyTorch~\cite{NEURIPS2019_9015} on Nvidia K80s GPUs;
the entire set of analyses can be computed within less than half a day on a single GPU.

\paragraph{Regression application I}
The first regression-type application  is the prediction of the unified Parkinson's disease rating scale (UPDRS) based on telemonitoring data.
The UPDRS is regularly used to assess the progression of the disease.
But measuring a UPDRS is expensive and inconvenient for patients,
because it requires time-consuming physical examinations by qualified medical personnel in a clinic.
A promising alternative to these direct, in person measurements is telemonitoring. 

The data~\cite{DuaUCI2019,tsanas_little_mcsharry_ramig_2010,little_mcsharry_hunter_spielman_ramig_2009,little_mcsharry_roberts_costello_moroz_2007} consists of $n=5875$ samples that each comprise a UPDRS score and $p=25$ associated patient attributes as discussed  in~\cite{tsanas_little_mcsharry_ramig_2010,little_mcsharry_hunter_spielman_ramig_2009}.
As usual, we standardize the data column-wise by subtracting the means and dividing through the standard deviations.

Our task is to predict the UPDRS of a new patient with attributes~$\boldsymbol{w}_1\in\mathbb{R}^{25}$ or of a group of new patients with attributes~$\boldsymbol{w}_1,\dots,\boldsymbol{w}_g\in\mathbb{R}^{25}$.
To test the performances of different approaches,
the inputs of samples (or groups of samples) selected uniformly at random are considered the ``new patient'' (or ``new group of patients''),
that is, they take the role of the  unlabeled target vector (or vectors) $\boldsymbol{w}_1$ (or $\boldsymbol{w}_1,\dots,\boldsymbol{w}_g$), 
and the corresponding outcome/outcomes are used for testing;
the remaining data is used as labeled training data $(\boldsymbol{x}_1,y_1),\dots ,(\boldsymbol{x}_n,y_n)$ (``previously telemonitored patients'').
We fit these data by a two-layer ReLU network with 150 and 50 neurons, respectively.
The training performances and the prediction performances on the targets are assessed by using the usual squared-error loss.

\begin{figure}[!h]
\centering
\begin{minipage}{.5\textwidth}
  \centering
  \includegraphics[width=0.72\linewidth]{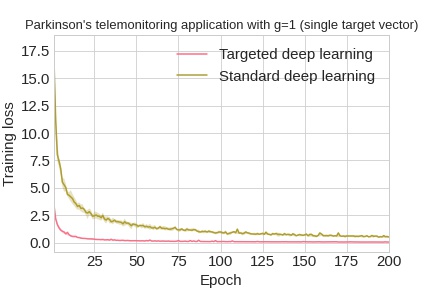}
\end{minipage}%
\begin{minipage}{.5\textwidth}
  \centering
  \includegraphics[width=0.72\linewidth]{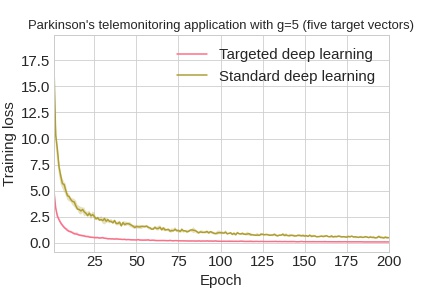}
\end{minipage}
\begin{minipage}{.5\textwidth}
  \centering
  \includegraphics[width=0.72\linewidth]{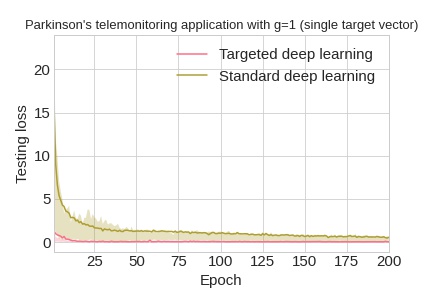}
\end{minipage}%
\begin{minipage}{.5\textwidth}
  \centering
  \includegraphics[width=0.72\linewidth]{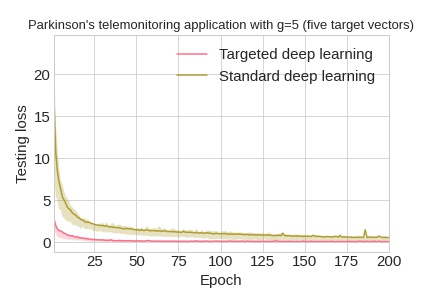}
\end{minipage}
\caption{training (top) and testing (bottom) losses for the Parkinson's telemonitoring data with single (left)/five (right) target vector/vectors.
Targeted deep learning yields faster and more accurate training than standard deep learning.}
\label{fig:parkinsons}
\end{figure}

\paragraph{Regression application II}
The second  regression-type application is the prediction of concrete strengths.
High-strength concrete is one of the most wide-spread building materials.
However, the production process of high-strength concrete is complicated:
it involves a large number of ingredients that can interact in intricate ways.
Hence, a mathematical model is very useful to predict the strength of a given mixture beforehand.

The data~\cite{DuaUCI2019,yeh_1998} consists of $n=1030$ samples that each comprises the compressive strength of a given mixture and $p=8$ associated attributes as discussed in~\cite{yeh_1998}.
The data are standardized as described above.

The task is to predict the compressive strengths of new mixtures of concrete.
Again, in line with our notion of targeted deep learning,
we assume given the target mixtures beforehand.
The performances are evaluated as described above.

\begin{figure}[!h]
\centering
\begin{minipage}{.5\textwidth}
  \centering
  \includegraphics[width=0.72\linewidth]{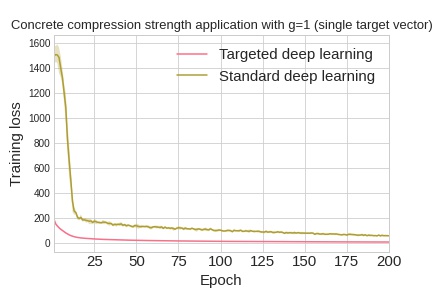}
\end{minipage}%
\begin{minipage}{.5\textwidth}
  \centering
  \includegraphics[width=0.72\linewidth]{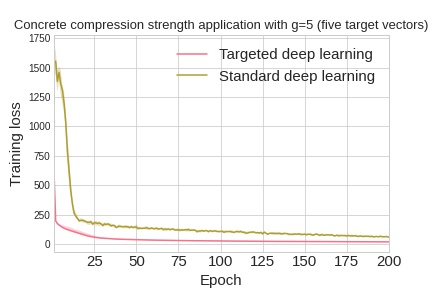}
\end{minipage}
\begin{minipage}{.5\textwidth}
  \centering
  \includegraphics[width=0.72\linewidth]{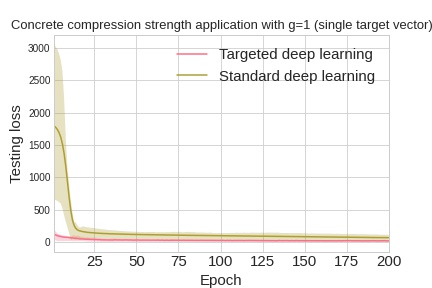}
\end{minipage}%
\begin{minipage}{.5\textwidth}
  \centering
  \includegraphics[width=0.72\linewidth]{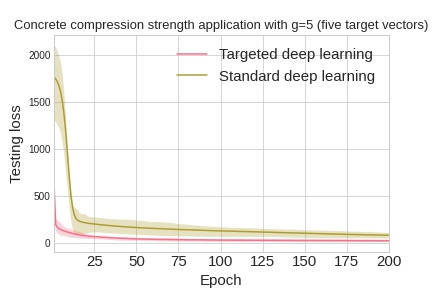}
\end{minipage}
\caption{training (top) and testing (bottom) losses for the concrete compression strength data with single (left)/five (right) target vector/vectors.
Targeted deep learning yields faster and more accurate training than standard deep learning.}
\label{fig:concrete}
\end{figure}

\paragraph{Classification application I}
The first  classification-type application is the detection of diabetic retinopathy.
Diabetic retinopathy is a medical condition of the eye caused by diabetes. 
It affects many long-term diabetes patients,
and it usually has only mild symptoms until it becomes a significant threat to the patient's vision. 
Thus, early diagnosis can improve healthcare considerably.

The data~\cite{DuaUCI2019,messiDR} consists of $n=1151$ samples extracted from the publicly accessible Messidor database of  patient fundus images~\cite{messiDR}.
Each sample comprises a binary observation of the existence of diabetic retinopathy and  $p=19$~image attributes as discussed in~\cite{antal_hajdu_2014}.
We standardize the features by subtracting the overall means and dividing through the overall standard deviations.

Our task is to classify a new/a group of patient in terms of diabetic retinopathy.
The data are fitted by a two-layer ReLU network with a cross-entropy output layer, where the layers have 150 and 50 neurons, respectively.
The training and testing accuracies are computed in terms of cross-entropy loss and classification error, respectively.

\begin{figure}[!h]
\centering
\begin{minipage}{.5\textwidth}
  \centering
  \includegraphics[width=0.72\linewidth]{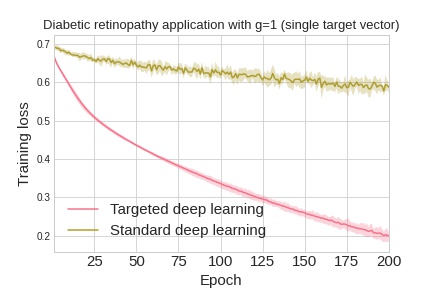}
\end{minipage}%
\begin{minipage}{.5\textwidth}
  \centering
  \includegraphics[width=0.72\linewidth]{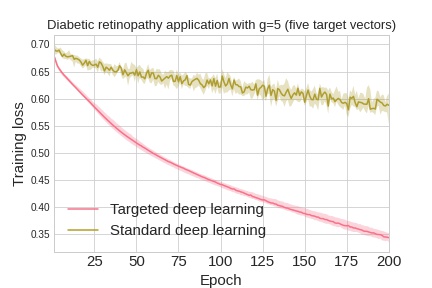}
\end{minipage}
\begin{minipage}{.5\textwidth}
  \centering
  \includegraphics[width=0.72\linewidth]{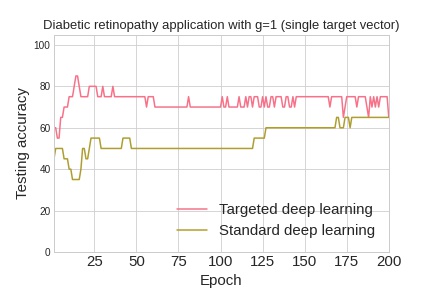}
\end{minipage}%
\begin{minipage}{.5\textwidth}
  \centering
  \includegraphics[width=0.72\linewidth]{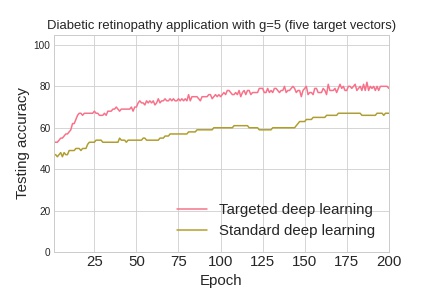}
\end{minipage}
\caption{training losses (top) and testing accuracies (bottom) for the diabetic retinopathy data with single (left)/five (right) target vector/vectors.
Targeted deep learning yields faster and more accurate training than standard deep learning.}
\label{fig:diabetes}
\end{figure}

\paragraph{Classification application II}
The second classification-type application is the classification of fetal cardiotocograms.
A cardiotocogram is a continuous electronic record that monitors the fetal heart rate and uterine contractions of a pregnant woman.
It is the most common diagnostic technique for evaluating the well-being of a fetus and its mother during pregnancy and right before delivery. 

The data~\cite{DuaUCI2019,ayres-de-campos_2000} consists $n=2126$~cardiotocograms of different pregnant women.
Each sample comprises a categorical quantity that denotes one of ten classes of fetal heart-rate patterns and 
$p=21$ cardiotocographic attributes as described in~\cite{ayres-de-campos_2000}.
The data are standardized as before.

The task is to classify (groups of) pregnant women regarding their fetal heart-rate patterns.
We test the performances at this task in the same way as in the diabetic retinopathy data set.

\begin{figure}[!h]
\centering
\begin{minipage}{.5\textwidth}
  \centering
  \includegraphics[width=0.72\linewidth]{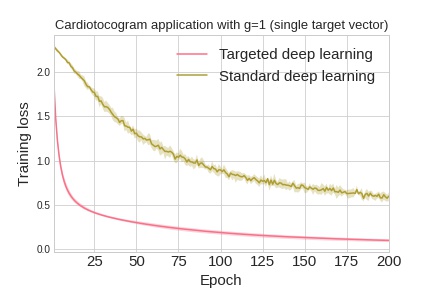}
\end{minipage}%
\begin{minipage}{.5\textwidth}
  \centering
  \includegraphics[width=0.72\linewidth]{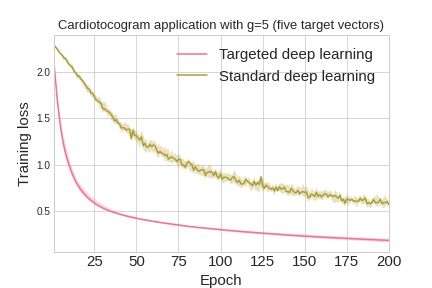}
\end{minipage}
\begin{minipage}{.5\textwidth}
  \centering
  \includegraphics[width=0.72\linewidth]{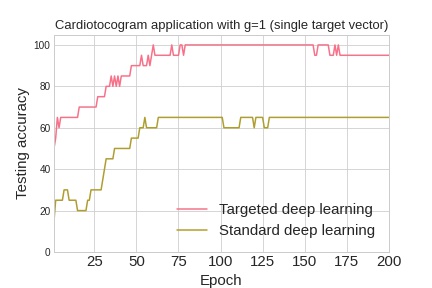}
\end{minipage}%
\begin{minipage}{.5\textwidth}
  \centering
  \includegraphics[width=0.72\linewidth]{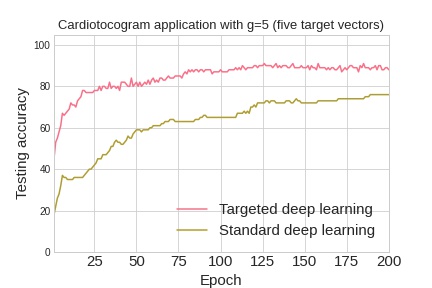}
\end{minipage}
\caption{training losses (top) and test accuracies (bottom) for the cardiotocogram data with single(left)/five (right) target vector/vectors.
Targeted deep learning yields faster and more accurate training than standard deep learning.}
\label{fig:card}
\end{figure}

\paragraph{Image classification I}
The first image application is the classification of handwritten digit images for the outputs within 10 different Japanese characters from the well-known Kuzushiji-MNIST (KMNIST) data set~\cite{rois-ds,kmnist}.

These data contain $n=60\,000$ samples for training, $10\,000$ samples for testing, where each sample is a $28 \time 28$-pixel grayscale image.
The data are standardized as described above.

Our task is to classify a new image.
The data are fitted by a two-layer ReLU convolutional neural network, where each convolutional layer is equipped with a max-pooling layer, and the kernel sizes for each convolution layer are 3 and 5, respectively.
We stress once more than we are not interested in the overall performance on the test data but instead in the performance on a (small) predefined set of images.
These performances are assessed as described above.

\begin{figure}[!h]
\centering
\begin{minipage}{.5\textwidth}
  \centering
  \includegraphics[width=0.72\linewidth]{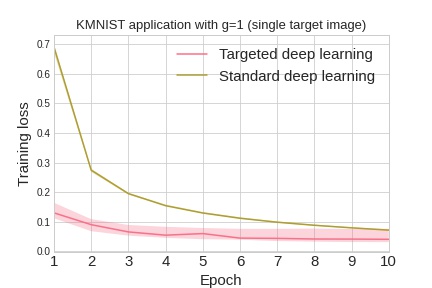}
\end{minipage}%
\begin{minipage}{.5\textwidth}
  \centering
  \includegraphics[width=0.72\linewidth]{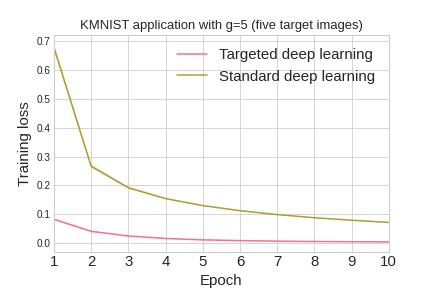}
\end{minipage}
\begin{minipage}{.5\textwidth}
  \centering
  \includegraphics[width=0.72\linewidth]{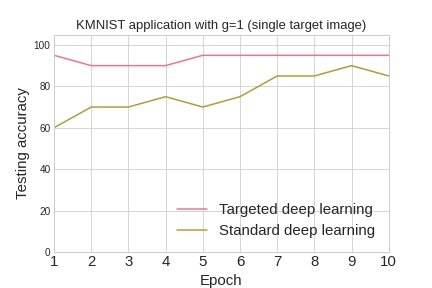}
\end{minipage}%
\begin{minipage}{.5\textwidth}
  \centering
  \includegraphics[width=0.72\linewidth]{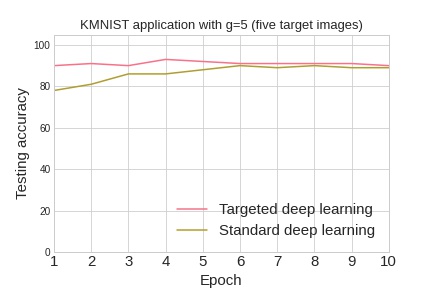}
\end{minipage}
\caption{training losses (top) and test accuracies (bottom) for the KMNIST data with single (left)/five (right) target image/images.
Targeted deep learning yields faster and more accurate training than standard deep learning.}
\label{fig:kmnist}
\end{figure}

\paragraph{Image classification II}
The second image application is the classification of popular item images with 10 labels from the well-known Fashion-MNIST data set~\cite{xiao2017/online}.

These data contain $n=60\,000$ samples for training, $10\,000$ samples for testing, where each sample is a $28 \time 28$-pixel grayscale image.
We proceed as in the previous application.

\begin{figure}[!h]
\centering
\begin{minipage}{.5\textwidth}
  \centering
  \includegraphics[width=0.72\linewidth]{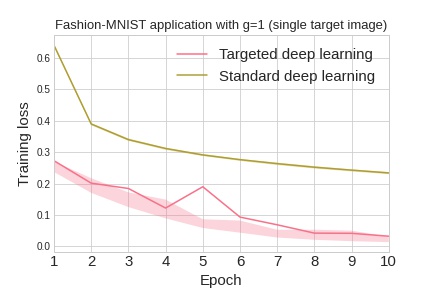}
\end{minipage}%
\begin{minipage}{.5\textwidth}
  \centering
  \includegraphics[width=0.72\linewidth]{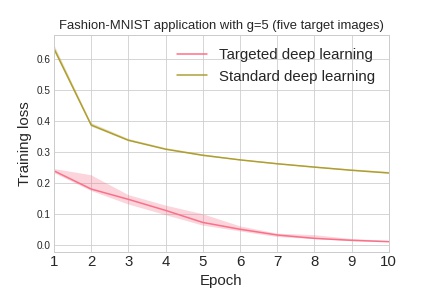}
\end{minipage}
\begin{minipage}{.5\textwidth}
  \centering
  \includegraphics[width=0.72\linewidth]{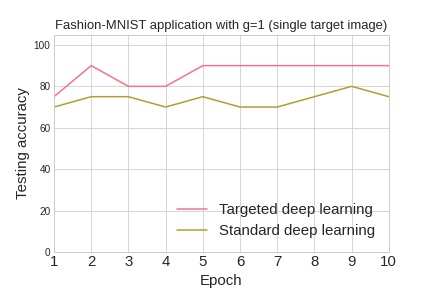}
\end{minipage}%
\begin{minipage}{.5\textwidth}
  \centering
  \includegraphics[width=0.72\linewidth]{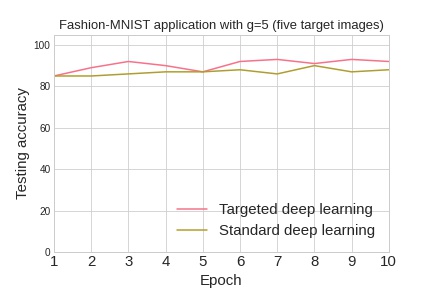}
\end{minipage}
\caption{training losses (top) and test accuracies (bottom) for the Fashion-MNIST data with single (left)/five (right) target image/images.
Targeted deep learning yields faster and more accurate training than standard deep learning.}
\label{fig:mnist}
\end{figure}

\paragraph{Results}
 Figures~\ref{fig:parkinsons}--\ref{fig:mnist} contain the results for individual targets and groups of $g=5$ targets.
The solid lines depict the means over 20~splits into training and testing samples and, where it makes sense,
the corresponding interquartile ranges as measures of uncertainty (recall that the  means can be outside of the interquartile ranges).
Across all applications, targeted deep learning renders the parameter training faster (indeed, the training and testing curves converge faster) and more accurate (testing curves converge to a higher level).

\paragraph{Conclusions}
The results indicate that  our framework and methods can gear network training very effectively to given target inputs.
Of course,
targeted deep learning only makes sense if the goal is indeed to optimize deep learning for certain inputs;
for other goals,  unsupervised learning, multimodal learning, personalized deep learning, or just standard deep learning might be more appropriate.
More generally speaking, 
it is important to select a suitable framework for a given task,
and targeted deep learning is a valuable addition to the set of possible frameworks.

\section{Discussion}
\label{sec:discussion}
Targeted deep learning can improve on standard deep learning when the goal is prediction or classification for target covariates that are known beforehand.
It can be implemented very easily either by adjusting the sampling strategy of SGD or by augmenting the training data.

Targeted deep learning is not restricted to specific types of architecture;
for example, it applies to convolutional neural networks equally well as to fully-connected neural networks.

Our approach already works very well with the simple correlation-based similarity measure discussed in Section~\ref{sec:methods}.
Resembling measures have been used in personalized deep learning~\cite{9078031}
and, of course, are common in statistics and machine learning more generally.
But our approach could also incorporate any other similarity measure,
including deep learning-based measures from the precision deep learning literature~\cite{8217759,che_xiao_liang_jin_zho_wang_2017}.

Targeted deep learning does not replace other paradigms,
such as minimizing overall generalization errors or personalized deep learning,
because it is sharply focused on situations where the  target covariates are known beforehand.
But if this is the case,
targeted deep learning can make a substantial difference,
leading, for example, to faster and more accurate predictions in medicine and engineering.

\ifarXiv
\paragraph{Acknowledgments}
We thank Mike Laszkiewicz, Yannick D\"uren, and Mahsa Taheri for their insightful suggestions.

\fi

\bibliographystyle{plain}
\bibliography{Literature}



\newpage
\appendix
\section{More on the sampling}
\label{sec:additional}
In this section,
we give some more insights into the sampling in our training algorithms described in Section~\ref{sec:methods}.
Recall first that our framework can be implemented by adjusting the batching algorithms directly,
which is the more elegant solution,
or by the augmentation scheme,
which is the more convenient solution.
The augmentation scheme involves a parameter~$t$,
but this parameter is of minor importance in practice.
We illustrate this in 
 Figure~\ref{fig:carda} below,
 which depicts the training losses and the testing accuracies on the cardiotocography data for a single target vector and for different choices of~$t$.
We find similar plots as in the corresponding figure in the main text,
showing that (i)~the data-augmentation scheme works as intended and (ii)~the exact choice of~$t$ is of minor importance.

Note also that neither approach adds new data (the data is sub-sampled from the existing data in either case), 
nor do they remove important information:
even for $t<1$,
important samples, that is, samples whose covariates are similar to at least one of the target covariates are included in the optimization with high probability.
Formulated differently,
the optimization might exclude samples,
but these samples are likely not informative for the task at hand.

\begin{figure}[!h]
\centering
\begin{minipage}{.5\textwidth}
  \centering
  \includegraphics[width=1.0\linewidth]{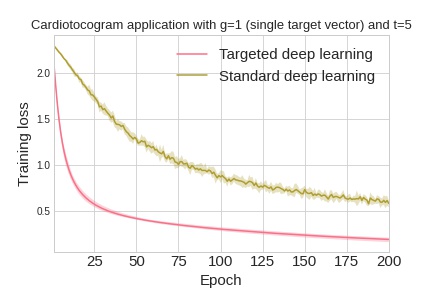}
\end{minipage}%
\begin{minipage}{.5\textwidth}
  \centering
  \includegraphics[width=1.0\linewidth]{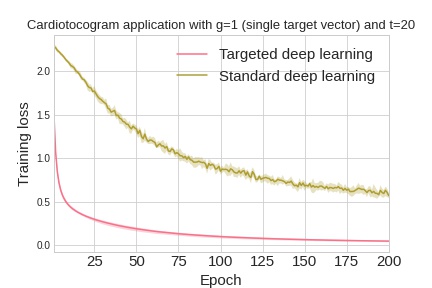}
\end{minipage}
\begin{minipage}{.5\textwidth}
  \centering
  \includegraphics[width=1.0\linewidth]{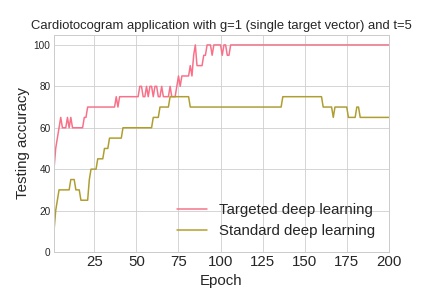}
\end{minipage}%
\begin{minipage}{.5\textwidth}
  \centering
  \includegraphics[width=1.0\linewidth]{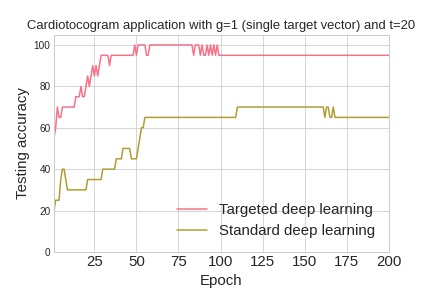}
\end{minipage}
\caption{
training losses (top) and test accuracies (bottom) for the cardiotocogram data with single target vector and $t \in \{5,20\}$
}
\label{fig:carda}
\end{figure}

\section{More on the similarity measure}
\label{sec:max}
In this section,
we provide further intuition about the similarity measure~$\mathfrak{s}$.
Observe first that $\mathfrak{s}[\boldsymbol{x},\boldsymbol{w}_1,\boldsymbol{w}_2,\dots]\in[0,1]$.
Moreover, it holds that 
\begin{equation*}
    \boldsymbol{x}~\text{is parallel to and in the same direction of at least one target vector}~\boldsymbol{w}_i~~\Leftrightarrow~~\mathfrak{s}[\boldsymbol{x},\boldsymbol{w}_1,\boldsymbol{w}_2,\dots]~=~1\,.
\end{equation*}
This makes sense:
if the covariate vector of a sample equals a target covariate vector,
the sample is highly informative and, therefore, should be included in the training with a high frequency.
On the other hand,
\begin{equation*}
    \boldsymbol{x}~\text{has an angle of at least 90 degrees to all}~\boldsymbol{w}_i~~~~~\Leftrightarrow~~~~~\mathfrak{s}[\boldsymbol{x},\boldsymbol{w}_1,\boldsymbol{w}_2,\dots]~=~0\,.
\end{equation*}
This also makes sense:
if the covariate vector of a sample is orthogonal to the target covariates or points in the opposite direction of the target covariates,
it has little in common with the target vectors and, therefore, can be excluded from the training.
This illustrates that the similarity measure,
especially the combination of the maximum and the angles,
can distinguish between relevant and irrelevant samples.

\section{More on the size of the target group}
\label{sec:group}

\begin{figure}[!h]
\centering
\begin{minipage}{.5\textwidth}
  \centering
  \includegraphics[width=1.0\linewidth]{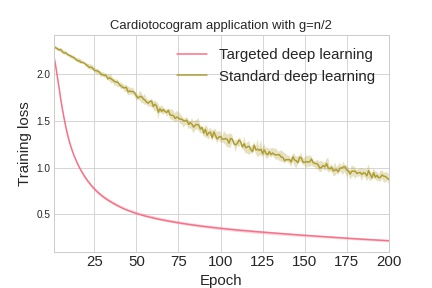}
\end{minipage}%
\begin{minipage}{.5\textwidth}
  \centering
  \includegraphics[width=1.0\linewidth]{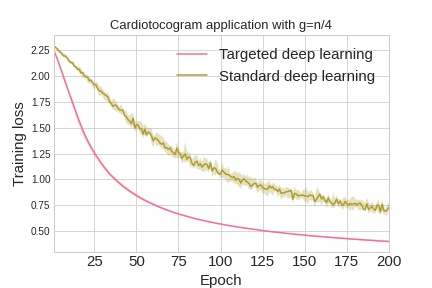}
\end{minipage}
\begin{minipage}{.5\textwidth}
  \centering
  \includegraphics[width=1.0\linewidth]{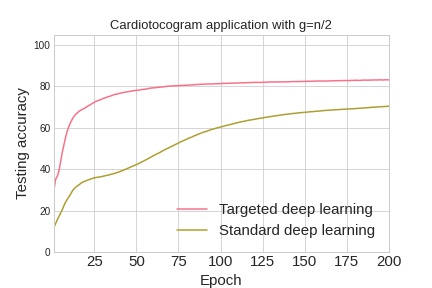}
\end{minipage}%
\begin{minipage}{.5\textwidth}
  \centering
  \includegraphics[width=1.0\linewidth]{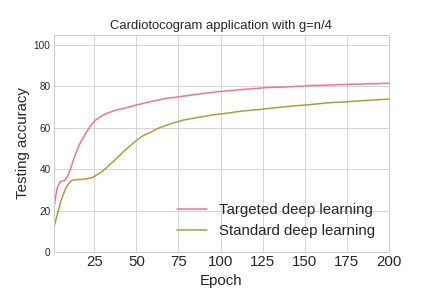}
\end{minipage}
\caption{
training losses (top) and testing accuracies (bottom) for the cardiotocogram data with $n/2$ (left) and $n/4$ (right) target vectors
}
\label{fig:group_carda}
\end{figure}

In this section, we illustrate that our approach also works for large groups of targets.
Figure~\ref{fig:group_carda} depicts the training losses and testing accuracies in the cardiotocography application for  $g=n/2$ and $g=n/4$.
The accuracies of our method declines as compared to the small-group cases, simply because fewer labeled samples are now available,
but the method still improves  the underlying standard deep-learning pipeline considerably.

\end{document}